\documentclass{ieeeaccess}
\usepackage{cite}
\usepackage{amsmath,amssymb,amsfonts}
\usepackage{amssymb}
\usepackage{pifont}
\usepackage{algorithmic}
\usepackage{graphicx}
\usepackage[small]{caption}
\usepackage{multirow}
\usepackage{textcomp}
\def\BibTeX{{\rm B\kern-.05em{\sc i\kern-.025em b}\kern-.08em
    T\kern-.1667em\lower.7ex\hbox{E}\kern-.125emX}}
\begin{document}
\history{Date of publication xxxx 00, 0000, date of current version xxxx 00, 0000.}
\doi{10.1109/ACCESS.2017.DOI}

\title{Dynamic Spatio-temporal Graph-based CNNs for Traffic Flow Prediction}
\author{
\uppercase{Ken Chen \authorrefmark{1}},
\uppercase{Fei Chen \authorrefmark{1}},
\uppercase{Baisheng Lai\authorrefmark{2}},
\uppercase{Zhongming Jin\authorrefmark{2}},
\uppercase{Yong Liu\authorrefmark{3}},
\uppercase{Kai Li \authorrefmark{1}},
\uppercase{Long Wei \authorrefmark{2}},
\uppercase{Pengfei Wang\authorrefmark{2}},
\uppercase{Yandong Tang \authorrefmark{1}},
\uppercase{Jianqiang Huang\authorrefmark{2}},
\uppercase{Xian-Sheng Hua\authorrefmark{2}}\IEEEmembership{Fellow, IEEE}
}

\address[1]{Sichuan Highway Transportation \& Communication Project Co., Ltd., Chengdu, China 
}
\address[2]{Alibaba Damo Academy, Alibaba Group, Hangzhou, China}
\address[3]{Alibaba Cloud, Alibaba Group, Hangzhou, China}
\markboth
{Somebody \headeretal: Dynamic Spatio-temporal Graph-based CNNs for Traffic Flow Prediction}
{Somebody \headeretal: Dynamic Spatio-temporal Graph-based CNNs for Traffic Flow Prediction}

\begin{abstract}

Forecasting future traffic flows from previous ones is a challenging problem because of their complex and dynamic nature of spatio-temporal structures. Most existing graph-based CNNs attempt to capture the static relations while largely neglecting the dynamics underlying sequential data. In this paper, we present dynamic spatio-temporal graph-based CNNs (DST-GCNNs) by learning expressive features to represent spatio-temporal structures and predict future traffic flows from surveillance video data. In particular, DST-GCNN is a two stream network. In the flow prediction stream, we present a novel graph-based spatio-temporal convolutional layer to extract features from a graph representation of traffic flows. Then several such layers are stacked together to predict future flows over time. Meanwhile, the relations between traffic flows in the graph are often time variant as the traffic condition changes over time. To capture the graph dynamics, we use the graph prediction stream to predict the dynamic graph structures, and the predicted structures are fed into the flow prediction stream. Experiments on real datasets demonstrate that the proposed model achieves competitive performances compared with the other state-of-the-art methods. 
\end{abstract}


\titlepgskip=-15pt

\maketitle

\section{Introduction}
\label{sec:introduction}
\PARstart{T}{he} goal of traffic flow forecasting is to predict the future traffic flows based on previous flows measured by sensors, which is one of the most challenging problems in Intelligent Transportation System (ITS). In the context of traffic flow forecasting, ``traffic flows'' or ``traffic volumes'' mean the number of cars recorded by a sensor network in a period of time. Accurate traffic forecasting can enable individuals and policy makers to make decisions on route planning and traffic control. 

%In the literature, this task has been studied for decades \cite{smith1997traffic, vlahogianni2004short‐term, liu2011discovering,lippi2013shortterm} but is still challenging because the spatio-temporal structure of traffic flows is complex and highly dynamic.

%To collect traffic flow data, multiple sensors include loop detectors \cite{jagadish2014big}, radar \cite{yu2017spatio} and GPS trajectories\cite{hoang2016fccf} can be leveraged to complement each other. For instance, GPS data may be incomplete since only a small portion of vehicles are equipped with GPS devices, although it can capture fine-grained trajectory flows of individual vehicles. Loop detector and radar can count the vehicles passing through them but only record coarse-grained flows at fixed checkpoints. In addition, we propose in this paper to use surveillance videos to record and predict traffic flows, which allows us to track many flows of vehicles simultaneously that appear in the camera views. Thanks to the fast development of computer vision technologies, we can not only track individual flows of vehicles but also link their identities across different cameras over time.

%\begin{figure}[t]
%  \centering
%  \includegraphics[width=0.45\textwidth]{./images/statements.png}
%  \caption{ Statements of traffic flow prediction task.}
%  \label{fig:statement}
%\end{figure}

Advanced algorithms that can model the interactions between dynamic traffic flows are required to predict their future trends. 
In literature, data-driven approaches have attracted many research attentions. For example, statistical methods such as autoregressive integrated moving average (ARIMA)\cite{davis1990adaptive} and its variants \cite{williams2003modeling} are well studied. The performance of such methods are limited because their capacity is  insufficient to model the complex nonlinear dependency among traffic flows in either spatial or temporal contexts. Recently, deep learning methods have shown promising results in dynamic prediction over sequential data, including stacked autoencoder (SAE) \cite{lv2015traffic}, DBN \cite{huang2014deep}, LSTM \cite{dai2017deeptrend} and CNN \cite{zhang2016deep}. Although these methods made some progress in modeling complex patterns in sequential data, they have not yet fully explored both spatial and temporal structures of traffic flows in an integrated fashion. 

Several methods \cite{li2017graph,yu2017spatio} attempt to model the traffic flows by unrolling static graphs through time where each vertex denotes the reading of flows at a given location and edges represent how the flows at two locations would affect each other. These works show that the graph structure is capable of describing the spatio-temporal dependency between flows. However, they usually have to assume that the graph structures, especially the relations between flows at different locations, do not change over time. It implies traffic conditions are time-invariant , which is not true in the real world.

To address this problem, we propose a dynamic spatio-temporal graph based CNN (DST-GCNN), which can model both the dynamics of traffic flows and their correlations. The contributions of this paper are threefold.
\begin{itemize}
\item We propose a novel spatio-temporal graph-based convolutional layer that is able to jointly extract both spatio and temporal information from the traffic flow data.
This layer consists of two factorized convolutions applied to spatial and temporal dimensions respectively, which significantly reduces computations and can be implemented in a parallel way. 
%is amenable to the implementations on the GPUs with many cores.
%It consists of two factorized convolutions applied to spatial and temporal dimensions respectively, 
%Compared with the LSTM layers, the proposed graph-based convolutional layer is much more efficient and amenable to the implementations on the GPUs with many cores, since it only consists of convolutional operations where the spatial and temporal convolutions are factorized.  
Then, we build a hierarchy of stacked graph-based convolutional layers to extract expressive features and make traffic flow predictions.

\item We will also learn the evolving graph structures that can adapt to the fast-changing traffic conditions over time. The learned graph structures can be seamlessly integrated with the stacked graph-based convolutional layers to make accurate traffic flow predictions.

\item We evaluate the proposed model on both traffic video dataset and the public Beijing taxi dataset. Experimental results demonstrate that DST-GCNN outperforms the state-of-the-art methods.

\end{itemize}

\section{Related Work}
The study of traffic flow forecasting can trace back to 1970s \cite{larry1995event}. From then on, a large number of methods have been proposed, and a recent survey paper comprehensively summarizes the methods \cite{vlahogianni2014short}. Early methods were often based on simulations, which were computationally demanding and required careful tuning of model parameters. With modern real-time traffic data collection systems, data-driven approaches have attracted more research attentions. In statistics, a family of autoregressive integrated moving average (ARIMA) models \cite{davis1990adaptive} 
%as well as their variants \cite{williams2003modeling,lippi2013shortterm} 
are proposed to predict traffic flows. However, these autoregressive models rely on the stationary assumption on sequential data, which fails to hold in real traffic conditions that vary over time. In \cite{hoang2016fccf}, Intrinsic Gaussian Markov Random Fields (IGMRF) are developed to model both the season flows and the trend flows, which is shown to be robust against noise and missing data. Some conventional learning methods including Linear SVR \cite{jin2007simultaneously} and random forest regression \cite{leshem2007traffic} have also been tailored to solve traffic flow prediction problem. Nevertheless, these shallow models depend on hand-crafted features and can not fully explore complex spatio-temporal patterns among the big traffic data, which greatly limits  their performances.

With the development of deep learning, various network architectures have been proposed for predicting traffic flows. Early attempts include SAE \cite{lv2015traffic} and DBN \cite{huang2014deep}, but neither are effective in modeling the spatio-temporal dependency between traffic flows. To capture the short and long temporal dependency, LSTMs are used to model the evolution of traffic flows  \cite{dai2017deeptrend}. However, the typical LSTM model is unable to model the spatial correlations which play an important role in making a spatially coherent prediction on the traffic flows. To close this gap, hybrid models where temporal models such as LSTM and GRU are combined with spatial models like 1D-CNN \cite{wu2016short} and graphs \cite{li2017graph} are proposed and achieve impressive performances. Nevertheless, recurrent models are restricted to process sequential data successively one-after-one,  which limits the parallelization of underlying computations. In contrast, the proposed model utilizes convolutions to capture both spatial and temporal data dependencies, which can reach much more efficiency than the compared recurrent models.

Two recent state-of-the-art methods are DCRNN \cite{li2017diffusion} and STGCN \cite{yu2017spatio}. They show appealing results on public datasets. But DCRNN is less efficient as it involves recurrent feedforward. STCGN use fixed affinity matrix that is not suitable for dynamic traffic environments. In contrast, our paper provides an efficient and dynamic method for better traffic prediction.

\section{Preliminaries}
\subsection{Structured Information Extraction}
\label{sec:structured}
To collect traffic flow data, multiple sensors including loop detectors \cite{jagadish2014big}, radar \cite{yu2017spatio} and GPS trajectories\cite{hoang2016fccf} can be leveraged. % to complement each other. 
However, they are either incomplete or unable to capture fine-grained trajectory flows of individual vehicles. 
%For instance, GPS data may be incomplete since only a small portion of vehicles are equipped with GPS devices, although it can capture fine-grained trajectory flows of individual vehicles. Loop detector and radar can count the vehicles passing through them but only record coarse-grained flows at fixed checkpoints. 
In contrast, we propose in this paper to use surveillance videos to record and predict traffic flows. %which allows us to track many flows of vehicles simultaneously that appear in the camera views. 
Thanks to the fast development of computer vision technologies, we can not only track individual flows of vehicles but also link their identities across different cameras over time.
%We first describe how to obtain the volumes of traffic flows and travel time from surveillance videos. 

In specific, we first use SSD \cite{liu2016ssd} and KCF \cite{henriques2015high} 
to identify and track vehicle instances in videos.
%for detecting and tracking vehicles. As a consequence, each individual vehicle instance is identified and tracked in videos. 
Then the plate numbers are recognized so that the same vehicles can be targeted across cameras. For each instance, we record the detected time, the camera location and its plate number, which are referred to as structured information in this paper. 
With them, at each location, we can count traffic volumes in a period of time. Meanwhile, by tracking plate numbers, we can estimate the average time to travel from one location to another. The computed traffic volumes and travel time are the inputs of our method.
%The computed traffic volumes and travel time are fed into the following parts of the framework.

It is worth noting that estimating the travel time across different locations does not require to recognize the plate numbers for all vehicles. The estimated travel time only need to be computed over those with recognized plate numbers, which greatly simplifies the problem.

\subsection{Mathematical Notations}
Suppose at each time $t$, we have volumes of traffic flows $\mathbf{F}_t\in \mathbb{R}^{C_0 \times N}$ and travel time $\mathbf{T}_t\in \mathbb{R}^{N\times N}$ data, where $N$ is number of locations and $C_0$ is the number of input channels. The channels represent different directions of traffic volumes at a location. 
%For example, if traffic flows are counted at cross roads, there are four channels of directions \{North, South, East, West\}. 
Our method uses $T_P$ previous traffic volumes $\{\mathbf{F}_{t-T_P+1},...,\mathbf{F}_t\}$ to forecast future volumes $\mathbf{F}_{t+T_F}$ after $T_F$ time steps. For simplicity, we use a tensor $\mathcal{X}_t\in\mathbb{R}^{C_0\times T_P\times N}$ to denote $\{\mathbf{F}_{t-T_P+1},...,\mathbf{F}_t\}$. Without ambiguity, the subscript $t$ may be ignored in the rest of paper.

%Now suppose we have traffic volumes $\{\mathcal{V}_t\in \mathbb{R}^{C \times N}, t=1,...,N_T\}$ and travel time $\{T_t\in \mathbb{R}^{N\times N}, t=1,...,N_T\}$ for $N_T$ time intervals, where $N$ is number of locations and $C$ is the number of channels. The channels represent different directions of traffic volumes at a location. For example, if traffic flows are counted at cross roads, there are four channels of directions \{North, South, East, West\}. At time $t$, $T_P$ previous traffic volumes $\{\mathcal{V}_{t-T_P+1},...,\mathcal{V}_t\}$ are used to forecast future traffic volumes at time $\mathcal{V}_{t+T_F}$. For simplicity, we use a tensor $\mathcal{X}_t\in\mathbb{R}^{C\times T_P\times N}$ to denote $\{\mathcal{V}_{t-T_P+1},...,\mathcal{V}_t\}$.

To model the complex dependency among traffic volumes, in DST-GCNN, we use an undirected graph $\mathbf{G}=(\mathbf{V}, \mathbf{A})$ to represent the traffic volumes, where the vertex set $\mathbf{V}$ represents traffic volumes at different locations and the affinity matrix $\mathbf{A}\in\mathbb{R}^{N\times N}$ depicts the connectivity between vertices. We derive the affinity matrix from the travel time such that $\mathbf{A}_{ij} = exp(-\mathbf{T}_{ij} /\sigma)$, where $\mathbf{T}_{ij}$ is the travel time between location $i$ and $j$. Therefore, the historical traffic volumes $\mathcal{X}_t$ can be represented as stacked graph frames.% as illustrated in Fig \ref{fig:statement}.

\section{Method}
\begin{figure*}[ht]
\centering
 \includegraphics[width=1.0\textwidth]{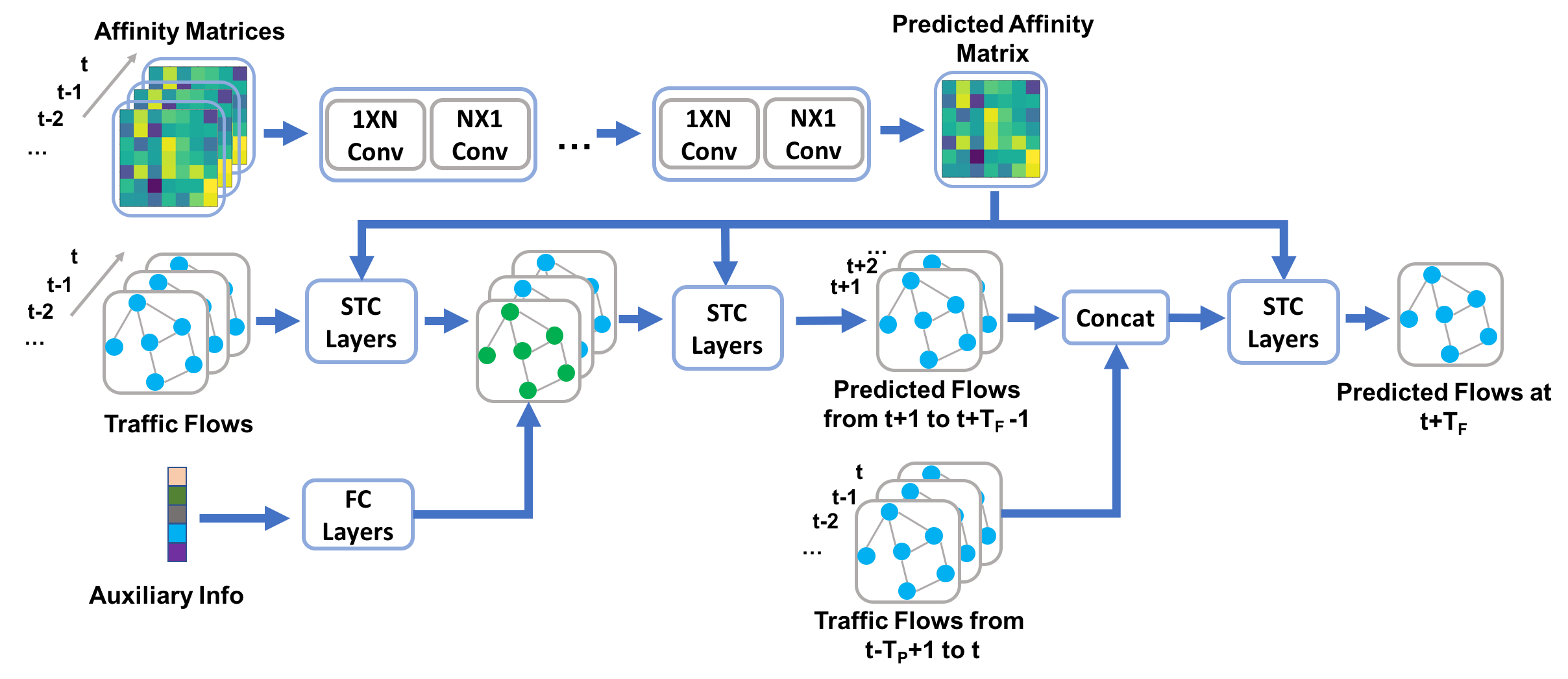}
 
  %\vspace{-0.2cm}
\caption{\footnotesize The overview of the proposed framework. The network consists of two streams, the first stream predicts the dynamic traffic conditions which are encoded in an affinity matrix. The second stream, equipped with the predicted traffic conditions and the proposed STC layers, first predicts future flows from $t+1$ to $t+T_F-1$, then predicts the target future flows at $t+T_F$.}
  \label{fig:overview}
\end{figure*}

The proposed DST-GCNN framework can model the complex spatio-temporal dependency between traffic flows and the fast-evolving traffic conditions. 
It takes three inputs: the previous traffic volumes represented as stacked graph frames, the previous traffic conditions represented as a series of affinity matrices and auxiliary information. Then these types of information are fed to a two-stream network. The graph prediction stream predicts the traffic conditions while the flow prediction stream forecasts evolutions of traffic flows given the predicted traffic conditions. The overall architecture of DST-GCNN is presented in Figure~\ref{fig:overview}. In the following subsections, we describe the two streams in details.

\subsection{Flow Prediction Stream}
In this subsection, we introduce the structure of the flow prediction stream that is the main sub-network to perform prediction. First, we present the building block of this stream, which is a novel Spatio-temporal Graph-based Convolutional Layer (STC) that works with spatio-temporal graph data. Then we build a two-step hierarchical model using STC layers to predict traffic flows.

%The building blocks of this path are the proposed Spatio-temporal Convolutional Layer (STC) 
\subsubsection{Spatio-temporal Graph-based Convolution}
\label{sec:sec_st}
The CNN is a popular tool in computer vision as it is powerful to extract hierarchy features expressive in many high-level recognition and prediction tasks. However, it cannot be directly applied to process the structured graph data like in our task. Therefore, we propose a novel layer that works with spatio-temporal graph data and is also as efficient as conventional convolutions. 

\begin{figure*}[t]
  \centering
  \includegraphics[width=0.7\textwidth]{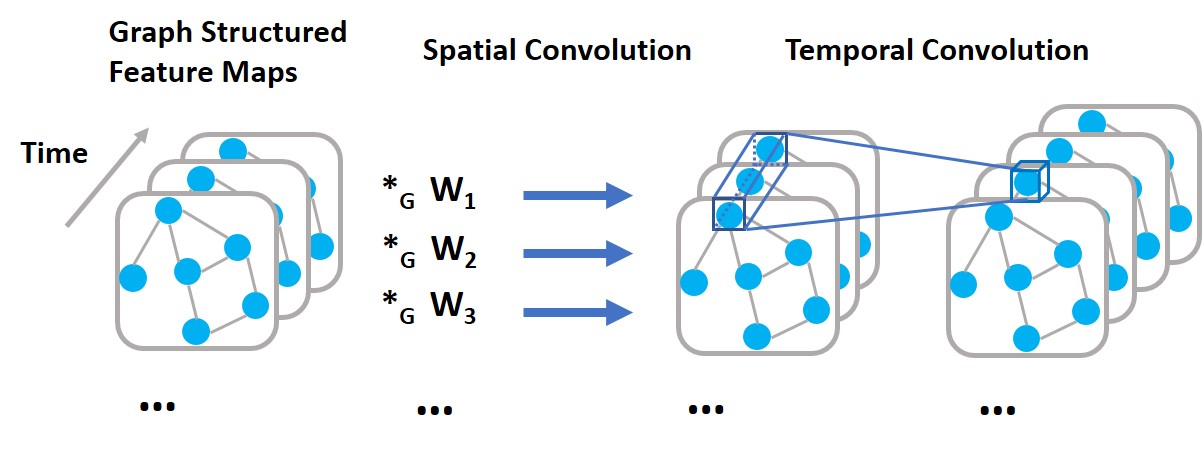}
  \caption{ The structure of spatio-temporal graph-based convolution. It consists of two convolutions that are applied on spatial and temporal dimensions respectively.}
  \vspace{-0.1cm}
  \label{fig:stc}
\end{figure*}

Inspired by \cite{howard2017mobilenets} that factorizes convolutions along two separate dimensions, we also present two factorized convolutions applied to spatial and temporal dimensions respectively, in a hope to reduce computational overhead. They form the proposed Spatio-temporal Graph-based Convolutional Layer (STC), whose structure is shown in Figure~\ref{fig:stc}. The input to a STC layer contains a sequence of graph structured feature maps organized by their timestamps and channels. Each graph is first convolved spatially to extract its spatial feature representation, and then features of multiple graphs are fused by a temporal convolution in a sliding time-window. In this way, both spatial and temporal information are merged to yield a dynamic feature representation for predicting future flows.

%\subsubsection{Spatial Convolution}
\vspace{0.2cm}
\noindent\textit{Spatial Convolution}
\vspace{0.05cm}

\noindent Let us define the spatial convolution on a given graph $\mathbf{G}=(\mathbf{V},\mathbf{A})$ first. The diagonal degree matrix and the graph Laplacian are defined as $\mathbf{D}_{ii} = \sum_{j}\mathbf{A}_{ij}$ and $\mathbf{L} = \mathbf{I} - \mathbf{D}^{-1/2} \mathbf{A} \mathbf{D}^{-1/2}$ respectively. Then the Singular Value Decomposition (SVD) is applied to Laplacian as $\mathbf{L} = \mathbf{U}\mathbf{\Lambda} \mathbf{U}^T$, where $\mathbf{U}$ consists of eigen vectors and $\mathbf{\Lambda}$ is a diagonal matrix of eigen values. The matrix $U$ is the Graph Fourier Transform matrix, which transforms an input graph signal $\mathbf{x} \in \mathbb{R}^{N}$ to its frequency domain  $\mathbf{Ux}\in \mathbb{R}^{N}$. With the same notation in \cite{henaff2015deep}, the convolution of a graph signal $\mathbf{x}$ with filter $\mathbf{g} \in\mathbb{R}^N$ on $\mathbf{G}$ is defined as
\begin{equation}
\label{eq:gc1}
\mathbf{x} *_{\mathbf{G}} \mathbf{g} = \mathbf{U}^{T}(\mathbf{Ug}\odot \mathbf{Ux}),
\end{equation}
where $\odot$ is the element-wise product. %The above equation uses the property that convolutions in spatial domain are equivalent to productions in frequency domain.

Let's define $\mathbf{w} = \mathbf{Ug}$ as the filter in frequency domain, then the convolution can be rewritten as 
%If the filter $\mathbf{g}$ is defined in frequency domain such that $\mathbf{w} = \mathbf{Ug}$,  the convolution can be rewritten as
\begin{equation}
\mathbf{x} *_{\mathbf{G}} \mathbf{g} =  \mathbf{U}^T (diag(\mathbf{w})\mathbf{Ux}),
\label{eq:gc2}
\end{equation}
%where the  filter $\mathbf{w}$ contains usually trainable parameters. 

The above graph convolution requires filter $\mathbf{w}$ to have the same size as input signal $\mathbf{x}$, which would be inefficient and hard to train when the graph has a large size. To make the filter ``localized'' as in CNN, $diag(\mathbf{w})$ can be approximated as polynomials of $\mathbf{\Lambda}$ \cite{defferrard2016convolutional} so that $diag(\mathbf{w}) = \sum_{k=0}^{K-1} \mathbf{\theta}_k \mathbf{\Lambda}^k$ 
%\begin{equation}
%diag(w) = \sum_{k=0}^{K-1} \theta_k \Lambda^k
%\end{equation}
and Eq \ref{eq:gc2} can be rewritten as
\begin{equation}
\mathbf{x} *_{\mathbf{G}} \mathbf{g} = \sum_{k=0}^{K-1} \mathbf{\theta}_k \mathbf{L}^k \mathbf{x}.
\end{equation}  

Now the trainable parameters become $\mathbf{\theta}\in\mathbb{R}^{K}$ whose size is restricted to $K$. In addition, a node is only supported by its $(K-1)$ neighbors \cite{hammond2011wavelets}.%the filter is $(K-1)$ localized on graph such as a node is only supported by its $(K-1)$ neighbors \cite{hammond2011wavelets}. 

Then we use the convolution operation above to define the spatial convolution in STC layer.
When computing the spatial convolution between feature map $\mathcal{X}^l \in \mathbb{R}^{C_l\times T_P\times N}$ and kernel $\mathcal{W}^l \in \mathbb{R}^{C_l\times T_P \times K}$ in the $l$-th layer of DST-GCNN, where $C_l$ is the channel number, the graph-based convolution defined above is applied to individual graph frame separately. In specific, each graph feature $\mathcal{X}^l_{c,p} \in \mathbb{R}^{N}$ at $c$-th channel and $p$-th time step is individually filtered such that
\begin{equation}
\label{sc}
\mathcal{Z}^l_{c,p} = \mathcal{X}^l_{c,p} *_{\mathbf{G}} \mathcal{W}^l_{c,p}, 
\end{equation}
where $\mathcal{W}^l_{c,p}\in \mathbb{R}^{K}$ and $\mathcal{Z}^l_{c,p}\in \mathbb{R}^{N}$ are the individual kernel and filtered output at the $c$-th channel and $p$-th time step, while tensor $\mathcal{Z}^l \in \mathbb{R}^{C_l \times T_P \times N}$ is the whole output.

%Denote $\mathcal{X}^l \in \mathbb{R}^{C_l\times T_P\times N}$ as feature map and $\mathcal{W}^l \in \mathbb{R}^{C_l\times T_P \times K}$ as the filter in the $l$-th layer, where $C_l$ is the channel number, 
%
%When computing the spatial convolution on feature map $\mathcal{X}^l \in \mathbb{R}^{C_l\times T_P\times N}$ in the $l$-th layer, where $C_l$ is channel number, the convolution defined above is applied to individual graph frame at each channel and time step separately such that
%\begin{equation}
%\label{sc}
%\mathcal{Z}^l_{c,p} = \mathcal{X}^l_{c,p} *_{\mathbf{G}} \mathcal{W}^l_{c,p},
%\end{equation}
%where tensor $\mathcal{Z}^l \in \mathbb{R}^{C_l \times T_P \times N}$ is the output, $\mathcal{W}^l \in \mathbb{R}^{C_l\times T_P \times K}$ is the filter. The subscripts $c$ and $p$ represent for the $c$-th channel and $p$-th time step.

%\subsubsection{Temporal Convolution}
\vspace{0.2cm}
\noindent\textit{Temporal Convolution}
\vspace{0.05cm}

\noindent At each time, after the spatial convolution, traffic flows are fused on the underlying graph, resulting in a multi-layered feature tensor $\mathcal{Z}^l$ compactly representing individual traffic flows and their spatial interactions. %modulated by the trainable parameters $W^l$ across different locations.

However, information across time steps is still isolated. To obtain spatio-temporal features, many previous methods \cite{jain2016structural,dai2017deeptrend,sutskever2014sequence} are based on recurrent models, which process sequential data iteratively step-by-step. Consequently, the information of current step is processed only when the information of all previous steps are done, which limits the efficiency of recurrent models. 

To make temporal operations as efficient as a convolution, we perform a conventional convolution along the time dimension to extract the temporal relations,  named after temporal convolution. For a feature tensor $\mathcal{Z}^l$ of size $[C_l, T_P, N]$, its convolution with kernel $\mathcal{K}^l$ of size $[C_l, C_{l+1}, Q, 1]$ is performed,
\begin{equation}
\label{tc}
\mathcal{X}^{l+1} = \mathcal{Z}^l * \mathcal{K}^l,
\end{equation}
where $Q$ is the size of time window. To keep the size of the time dimension unchanged, we pad $(Q-1)/2$ zeros on both sides of the time dimension.
%the second time dimension of $\mathcal{Z}^l$ before the temporal convolution.

%\subsubsection{Putting Together: Spatio-Temporal Graph-based Convolution }
\vspace{0.2cm}
\noindent\textit{Putting Together}
\vspace{0.05cm}

\noindent By combining Eq. \ref{sc} and Eq. \ref{tc}, we have the following definition of spatio-temporal graph-based convolution:
\begin{equation}
\mathcal{X}^{l+1} = STC(\mathcal{X}^l, \mathcal{W}^l, \mathcal{K}^l, \mathbf{G}),
\end{equation}
whose structure is shown in Figure \ref{fig:stc}.
%which is named after STC layer in the rest of this paper and the structure is shown in Figure \ref{fig:stc}. %To introduce non-linearity in STC layer, a ReLU layer is inserted between spatial and temporal convolutions. 

We now analyse the efficiency of our factorized convolution. Without such factorization, one needs to build a graph with $N\times C_l\times T_P$ nodes to capture both spatial and temporal structures, making the graph convolution in Eq. \ref{eq:gc2} have complexity of $\mathcal{O}(N^2C_l^2T_P^2)$. While our STC layer builds $C_l \times T_P$ graphs with $N$ nodes and separates spatial and temporal convolutions, has complexity of $\mathcal{O}(N^2 C_l T_P+N C_l C_{l+1} T_P Q)$, which is much more efficient.

\subsubsection{Two-step Prediction}
\label{sec:twostep}
The STC layers are able to jointly extract both spatial and temporal information from the sequence of traffic flows. We can build a hierarchical model using such layers to extract features and predict future flows from previous flows $\{\mathbf{F}_{t-T_P+1},...,\mathbf{F}_{t}\}$. 
%Using the proposed STC layer as building block, we can build a hierarchical model to extract features and predict future flows from previous flows $\{\mathbf{F}_{t-T_P+1},...,\mathbf{F}_{t}\}$. 
A straight way is to directly predict future traffic volume $\mathbf{F}_{T_F}$ after $T_F$ intervals as existing methods \cite{dai2017deeptrend,yu2017spatio,defferrard2016convolutional}. This one-step prediction scheme is simple but has two disadvantages. First, it only uses ground truth data at $t+T_F$ to train the model but neglects those between $t+1$ and $t+T_F-1$. Second, when $T_F$ is large, it is hard for one-step methods to capture traffic trends for such a long time, since the input and the future volumes may be very different.

%Given previous volumes of traffic flows $\{\mathbf{F}_{t-T_P+1},...,\mathbf{F}_{t}\}$, existing traffic prediction methods \cite{dai2017deeptrend,yu2017spatio,defferrard2016convolutional} directly predict future traffic volume $\mathbf{F}_{T_F}$ after $T_F$ intervals. This one-step prediction scheme is simple but has two disadvantages. First, it only uses ground truth data at $t+T_F$ to train the model but neglects those between $t+1$ and $t+T_F-1$ which could be useful to increase the prediction accuracy. Second, when $T_F$ is large, it is hard for one-step methods to capture traffic trends for such a long time, since the input and the future volumes may be very different.

%Second, when $T_F$ is large, the gap between the input volumes and the prediction volumes can be large, 
%making the one step prediction ineffective because the methods must capture trends for a long time.

To solve the above issues, we propose a new prediction scheme that divides the prediction problem into two steps. In the first step, we use previous flows $\{\mathbf{F}_{t-T_P+1},...,\mathbf{F}_{t}\}$ to predict future volumes between $t+1$ and $t+T_f-1$, which are called ``close future flows''. During the training phase, the predicted ``close future flows'' are supervised by ground truth at the corresponding time period. As a result, ground truth data between $t+1$ and $t+T_F-1$ is imposed into training procedure. In the second step, the ``target future flows'' at time $t+T_F$ is predicted by considering both previous flows and the predicted ``close future flows''. Compared with one-step methods, the prediction of ``target future flows'' is easier now since it utilizes ``close future flows'' and it only predicts one step further. The two-step prediction scheme is shown in the second path in Figure ~\ref{fig:overview}.

%which is denoted as $\mathcal{Y}_t = \{\mathcal{V}_{t+1},...,\mathcal{V}_{t+T_f-1}\}$ and

Let's denote the models of the first step and the second step as $\mathcal{M}_{S1}$ and $\mathcal{M}_{S2}$ respectively. These two model both stacks several STC layers for prediction. The loss function of two-step prediction can be written as:
\begin{equation}
L_{two-step} = \|\mathcal{Y}_t - \hat{\mathcal{Y}}_t\|^2 + \|\mathbf{F}_{t+T_F} - \mathcal{M}_{S2}(\mathcal{X}_t,\hat{\mathcal{Y}}_t,\mathbf{\Theta}_{S2})\|^2,
\label{eq:loss_two}
\end{equation}
where $\hat{\mathcal{Y}}_t = \mathcal{M}_{S1}(\mathcal{X}_t, \mathbf{\Theta}_{S1})$ is the predicted ``close future flows'' and $\mathcal{Y}_t = \{\mathbf{F}_{t+1},...,\mathbf{F}_{t+T_F-1}\}$ is the ground truth.  $\mathbf{\Theta}_{S1}$ and $\mathbf{\Theta}_{S2}$ are parameters of two models respectively.

%Like the CNNs, the hierarchical baseline model extracts multi-level features.The bottom layers with smaller spatio-temporal extents represent low-level features, which perceive local and short-time traffic trends. Then the middle layers with larger spatio-temporal extents extract traffic trends of larger regions for a longer period. Finally, the top layers model the most complex patterns of global traffic trends, which can predict the traffic volumes into deep future. A noticeable difference from the CNNs is the absence of pooling layers so that high-level features will not miss any fine-grained spatio-temporal information. 

%Denote the baseline model by $\mathcal{M}_{baseline}$, a MSE (mean square error) loss is minimized to model it:
%\begin{equation}
%L_{baseline} = \frac{1}{N_T} \sum_t\| \mathcal{M}_{baseline}(\mathcal{X}_t, \Theta, \mathbf{G}_B) - \mathbf{F}_{t+T_F} \|^2,
%\end{equation}
%where $\mathbf{\Theta}$ denotes the trainable parameters. 

\subsubsection{Auxiliary Information Embedding}
\label{sec:auxiliary}
Except for previous flows, some auxiliary information like time, the day of week and weather are useful to predict future flows. The influence of such information is studied in \cite{hoang2016fccf,zhang2016deep}. For example, weekdays and weekends have very different transit patterns and a thunder storm can suddenly reduce the traffic volumes.

To make full use of such auxiliary information, we embed them into the traffic flow prediction network. We first encode these information into one-hot vectors. %For example, we encode time into a 48 length one-hot vector, which represents the index of half hour in the day. The day of week is encoded into a vector of length 7. 
Then these one-hot vectors are concatenated and we use several fully connected layers to extract a feature vector. The feature vector is later reshaped so that it can be concatenated with traffic flow feature maps. Finally, the concatenated features are fed into prediction modules, as shown in Figure ~\ref{fig:overview}.

\subsection{Graph Prediction Stream}
%\subsection{Dynamic Graph Learning}
\label{sec:dynamic}
In this subsection, we introduce the other stream in the framework, which is named as the graph prediction stream. Previous methods \cite{henaff2015deep,jain2016structural,yu2017spatio} that model spatio-temporal graphs assume that the graph structure of spatio-temporal data is fixed without temporal evolutions. However, in real world applications, the graph structures are dynamic. For instance, in the traffic prediction problem, traffic conditions are time-variant, implying that the connectivities between vertices in graphs change over time. In order to model such dynamics, we introduce a stream in the framework to predict such time-variant graph structures.

In particular, at each time $t$, we have a graph structure $\mathbf{G}_t$ for STC layers in the model as a function of time $t$. It reflects the average traffic condition in the period between time $t-T_P+1$ and $t+T_F$. 
%One way to obtain $G_t$ is first computing the average travel time in the corresponding period
%\begin{equation}
%\bar{T}_{t-T_p:t+T_f} = \frac{1}{T_f + T_p +1}\sum_{i=t-T_p}^{t+T_f} T_i.
%\end{equation}
%Then we have the average affinity matrix $\bar{A}_{t-T_p:t+T_f}$ and the corresponding Laplacian. 
%One way to obtain $G_t$ is first computing the average travel time in the corresponding period and then compute the Laplacian as we do in the baseline model. 
$\mathbf{G}_t$ can not be directly computed since the future travel time during $t+1$ to $t+T_F$ is unavailable in the test phase. 
%In the test phase, the future travel time during $t+1$ to $t+T_F$ is unavailable. 
To address this problem, we introduce another path in the network to predict graph structure $\mathbf{G}_t$ from previous travel time data $\{\mathbf{T}_{t-T_P+1},...,\mathbf{T}_{t}\}$. 
%In other words, we predict the average traffic condition during $t-T_P+1$ to $t+T_f$ using previous data from $t-T_P+1$ to $t$. 
Specifically, $\{\mathbf{T}_{t-T_P+1},...,\mathbf{T}_{t}\}$ are first converted to affinity matrices to construct a tensor $\mathcal{S}_t = \{\mathbf{A}_{t-T_P+1},...,\mathbf{A}_{t}\}\in \mathbb{R}^{T_P\times N\times N}$, then it is fed into a sub-network $\mathcal{M}_{G}$ to predict a new affinity matrix $\hat{\mathbf{A}}_t=\mathcal{M}_{G}(\mathcal{S}_t, \mathbf{\Theta}_{G})$ representing for the average traffic condition during $t-T_P+1$ and $t+T_F$, where $\mathbf{\Theta}_{G}$ is parameter of $\mathcal{M}_{G}$.

%Specifically, the travel time data at each $t$ are converted to affinity matrix $A_t = exp(-T_t / \sigma)$. Then a sub-network is used to predict an affinity matrix $\hat{A}_{t-T_P+1:t+T_f}$ from previous ones $\mathcal{A}_t = \{A_{t-T_p},...,A_{t}\}\in \mathbb{R}^{T_p+1\times N\times N}$. 
During training, the graph prediction stream is supervised by minimizing the following loss function
%\begin{small}
\begin{equation}
L_{dynamic} = \sum_t \|\hat{\mathbf{A}}_t - \bar{\mathbf{A}}_t \|_1,
\label{eq:loss_dy}
\end{equation}
%\end{small}
where $\bar{\mathbf{A}}_t \in \mathbb{R}^{N\times N}$ is the ground truth average affinity matrix during $t-T_P+1$ and $t+T_F$. $L_1$ norm is used to avoid the loss from being dominated by some large errors. The Laplacian of $\hat{\mathbf{A}}_t$ is then computed and fed into STC layers. In this way, the prediction model takes the dynamic traffic conditions into consideration, thus it is able to make more accurate predictions on future traffic flows.

To model the relations of previous affinity matrices, a model with global field of view is required since entries of affinity matrices have ``global'' correlations. For instances, $\mathbf{A}_{ij}$ and $\mathbf{A}_{ji}$ is closely related no matter how apart they are located in $\mathbf{A}$. Thus, we stack multiple pairs of convolutional layers, where each pair consists of convolutional layers of kernel sizes $[1,N]$ and $[N,1]$ respectively to get the large spatial extent, as shown in the first stream of Figure ~\ref{fig:overview}.

%For the graph structure prediction network, stacked fully connected layers may be preferred because entries of affinity matrices have ``global'' correlations. For instances, $A_{ij}$ and $A_{ji}$ is closely related no matter how apart they are located in $A$. However, fully connected layers are hard to train because they have a large number of parameters. In addition, affinity matrices are sparse, which makes many parameters in fully connected layers redundant. 

%To handle this issue, we use convolutional layers instead of fully connected layers. In particular, multiple pairs of convolutional layers are stacked, where  each pair consists of convolutional layers of kernel sizes $[1,N]$ and $[N,1]$ respectively to get the large spatial extent. Here $N$ is the number of rows and columns of affinity matrices. In our experiment, such convolutional layers achieve better performance than fully connected layers. 

\subsection{The Whole Model}
By combining the two paths, we get the full model of DST-GCNN shown in Figure ~\ref{fig:overview}. The loss function of the complete model is 
\begin{equation}
	L = L_{two-step} + L_{dynamic}
\end{equation}

%The network is trained by two losses, one is for dynamic graph learning defined in Eq \ref{eq:loss_dy}, the other is for traffic flow prediction as defined in Eq \ref{eq:loss_two}. 

It is worth noting that DST-GCNN is a general method to extract features on spatio-temporal graph structured data, it can be  applied to not only traffic flow prediction tasks, but also other more general regression or classification tasks on graph data, especially when the graph structure is dynamic. For instance, it can be adapted to skeleton-based action recognition or pose forecasting tasks with minor modification.

%\subsection{The Baseline Model}
%\label{sec:baseline}
%The STC layers are able to jointly extract both spatial and temporal information from the sequence of traffic flows. To verify the effectiveness of STC layers in modeling such spatio-temporal graph data, we build a baseline model that stacks $L$ STC layers and a ReLU layer to predict future traffic volumes. The ReLU layer ensures non-negative outputs.
%
%For simplicity, the baseline model assumes the same graph structure all the time. Later we will abandon this assumption by allowing the changing structures over time. The static graph structure, denoted by $\mathbf{G}_B$, can be computed by averaging the whole set of travel times $\{\mathbf{T}_t, t=1,...,N_T\}$, where $N_T$ is the number of time intervals. In particular, the affinity matrix $\mathbf{A}_B$ of $\mathbf{G}_B$ is first computed by $\mathbf{A}_B = exp(-\mathbf{T}_B/\sigma)$ where $\mathbf{T}_B$ is the average travel time, and then the Laplacian $\mathbf{L}_B$ are calculated as described in \ref{sec:sec_st}.

\section{Experiments}

In this section, we present a series of experiments to assess the performance of the proposed methods.
We first introduce the datasets and the implementation details of DST-GCNN. Then we conduct ablation experiments to evaluate the effectiveness of components in DST-GCNN. At last, our method are compared with state-of-the-art methods on these datasets.
 
\subsection{Dataset and Evaluation Metrics}
\label{sec:dataset}
Our experiments are conducted on two public datasets: METR-LA \cite{jagadish2014big} and TaxiBJ \cite{hoang2016fccf}, and our collected dataset CD-HW.

We first introduce two public available \textbf{METR-LA} \cite{jagadish2014big} and \textbf{TaxiBJ} \cite{hoang2016fccf}.
METR-LA is a large-scale dataset collected from 1500 traffic loop detectors in Los Angeles country road network. This dataset includes speed, volume and occupancy data, covering approximately 3,420 miles. As \cite{li2017diffusion}, we choose four months of traffic speed data from Mar 1st 2012 to Jun 30th 2012 recorded by 207 sensors for our experiment. The traffic data are aggregated every 5 minutes with one direction.
%The traffic volumes and travel time data of 
TaxiBJ Dataset are obtained from taxis' GPS trajectories in Beijing during 1st March to 30th June 2015. The authors partition Beijing into 26 high-level regions and traffic volumes are aggregated in every 30 minutes, with two directions \{In, Out\}. Besides crowd flows, it also includes weather conditions that are categorized into good weather (sunny, cloudy) and bad weather (rainy, storm, dusty). 

\begin{figure}[th]
  \centering
  \includegraphics[width=0.40\textwidth]{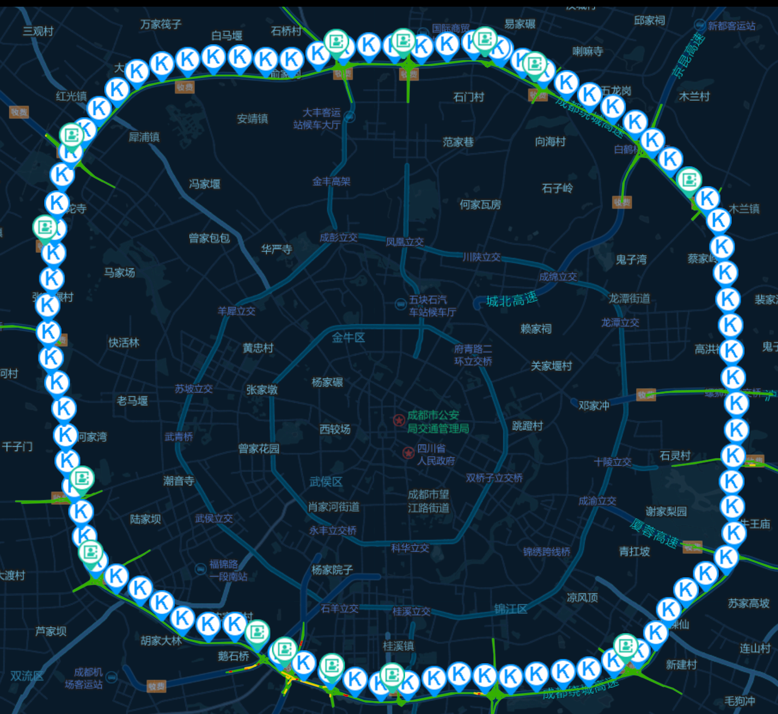}
  \caption{Distribution of sensors on highway around Chengdu City in the CD-HW dataset. }
  \label{fig:map}
\end{figure}

We also collect a new dataset that contains speed data recorded along highway around Chengdu city, China.
This dataset is named as Chengdu Highway (\textbf{CD-HW}) Dataset. CD-HW dataset is capured by 1,692 roadside sensors during 1 Nov to 30 Nov 2019. Their locations are shown in Figure \ref{fig:map}.
%We select videos between 6am to 12pm since there are almost no vehicles before 6am. 
%Traffic flows at each cross road and the travel time between adjacent cross roads are obtained as described in \ref{sec:structured}. 
%Auxiliary information including time and day of week is provided in this dataset. 
Data before 25th Nov are used for training and the remaining for test. The speed data are colleced every 10 minutes with one direction.
%In experiments, traffic volumes and travel time are distilled from videos by object detection and tracking algorithms introduced in subsection ~\ref{sec:structured}, and the results are aggregated in an interval of every eight minutes. 
%Then traffic flows of the last eight intervals are used to predict those after five intervals.

For evaluation, we use the Root Mean Squared Error (RMSE) metric, the Mean Absolute Percentage Error (MAPE) and the Mean Absolute Error (MAE), which are defined as below:
% and the Mean Average Error (MAE)
%\begin{small}
\begin{equation}
\begin{aligned}
RMSE &= \frac{1}{N_T}\sum_{t=1}^{N_T}\sqrt{\frac{1}{N}\sum_{i=1}^{N}(\hat{\mathbf{F}}_{t,i} - \mathbf{F}_{t,i})^2}, \\
MAPE &= \frac{1}{N_T}\sum_{t=1}^{N_T}  \frac{1}{N}\sum_{i=1}^{N}|\frac{\hat{\mathbf{F}}_{t,i} - \mathbf{F}_{t,i}}{\mathbf{F}_{t,i}}|, \\
MAE &= \frac{1}{N_T}\sum_{t=1}^{N_T}  \frac{1}{N}\sum_{i=1}^{N}|\hat{\mathbf{F}}_{t,i} - \mathbf{F}_{t,i}|,
\end{aligned}
\end{equation}

where $\hat{\mathbf{F}}_{t,i}$ and $\mathbf{F}_{t,i}$ are the predicted and ground truth traffic volumes (speed) at time $t$ and location $i$.

\subsection{Implementation Details}
Models $\mathcal{M}_{S1}$ and $\mathcal{M}_{S2}$ presented in subsection \ref{sec:twostep} consist of three STC layers with $8$, $16$, $32$ channels respectively. A ReLU layer is inserted between two STC layers to introduce nonlinearity as CNNs. Another ReLU layer is added after the last STC layer to ensure non-negative prediction. In spatial convolution of STC layer, the order $K$ of polynomial approximation is set to be 5 and the temporal convolution kernel size is set to be $5\times 1$. The graph prediction stream $\mathcal{M}_{G}$ consists of three pairs of $1\times N$ and $N \times 1$ convolutional layers with $16$ channels. The auxiliary information is encoded by two fully connected layers with $32$ and $N\times N \times C \times T_p$ output neurons respectively, so that the output can be reshaped and concatenated with flow features. In the training procedure, we first pre-train the dynamic graph learning sub-network for $10$ epochs and jointly train the whole model for $100$ epochs. The model is trained by SGD with momentum. The first 50 epochs take a learning rate of $10^{-2}$ and the last 50 epochs use $10^{-3}$. Finally, the framework is implemented by PyTorch.
\begin{table*}[!th]
\centering
	\renewcommand{\arraystretch}{1.1}
\begin{tabular} {l|c|c|c|c|c|c}
\hline
\multirow{2}{*}{Method} & \multicolumn{3}{|c|}{Out Volumes} & \multicolumn{3}{|c}{In Volumes} \\
\cline{2-7}
 & MAE & RMSE & MAPE & MAE & RMSE & MAPE \\
\hline
Basel & 10.49 & 13.48 & 13.11\% & 10.71 & 14.44 & 14.46\%  \\
%\hline
Basel+AE & 10.24 & 13.18 & 12.81\% & 10.41 & 13.96 & 14.35\%  \\
%\hline
Basel+AE+GP & 10.03 & 12.88 & 12.75\% & 10.40 & 13.94 & 14.39\%  \\
%\hline
DST-GCNN & \textbf{9.93} & \textbf{12.78} & \textbf{12.56\%} & \textbf{10.24}& \textbf{13.78} & \textbf{14.02\%} \\
\hline
\end{tabular}
%\vspace{0.2cm}
\caption{Performance comparison of our models with different configurations on TaxiBJ dataset.}
\label{ablation}
\end{table*}

\begin{table*}[thp]
	\centering
	\renewcommand{\arraystretch}{1.1}
	\caption{Performance comparison of different methods on the METR-LA dataset. MAE, RMSE and MAPE(\%) metrics are compared for different predicting horizons. Bold numbers indicate the best results.}
	\begin{tabular}{cccccccc}  
		\hline
		\multicolumn{1}{c}{T} & \multicolumn{1}{c}{Metric} & \multicolumn{1}{c}{ARIMA}   & \multicolumn{1}{c}{FNN} & \multicolumn{1}{c}{FC-LSTM} & \multicolumn{1}{c}{DCRNN} & \multicolumn{1}{c}{STGCN} &
		\multicolumn{1}{c}{DST-GCNN} \\
		\hline
		       & MAE  & 3.99  & 3.99 & 3.44 & 2.77 & 2.87 & \textbf{2.68}\\
		15 min & RMSE & 8.21  & 7.94 & 6.30 & 5.38 & 5.54 & \textbf{5.35}\\
		       & MAPE & 9.6    & 9.9  & 9.6  & 7.3  & 7.4 & \textbf{7.2} \\ 
		\cline{2-8}
		       & MAE  & 5.15 & 4.23 & 3.77 & 3.15 & 3.48 & \textbf{3.01} \\
		30 min & RMSE & 10.45  & 8.17 & 7.23 & 6.45 & 6.84 & \textbf{6.23} \\
		       & MAPE & 12.7 & 12.9 & 10.9 & 8.8 & 9.4 & \textbf{8.5} \\ 
		\cline{2-8}
		 	   & MAE  & 6.90 & 4.49 & 4.37 & 3.60 & 4.45 & \textbf{3.41} \\
		60 min & RMSE & 13.23  &  8.69 & 8.69 & 7.59 & 8.41& \textbf{7.47} \\
			   & MAPE & 17.4  & 14.0 & 13.2 & 10.5 & 11.8 & \textbf{10.3} \\ 
		\hline
	\end{tabular} 
	\label{tab:LA_arts}
\end{table*}
\subsection{Ablation Study}
\label{sec:ablation}
To investigate the effectiveness of each component, we first build a plain \textbf{baseline model} which stacks three STC layers as $\mathcal{M}_{S1}$ while uses one-step prediction scheme, keeps graph structure fixed and does not use auxiliary information. The static graph structure is calculated by averaging all traffic time in training set.
Then different configurations are tested, including: 
\begin{itemize}
\item The baseline model denoted as denoted as \textbf{Basel};
\item The baseline model with auxiliary information embedding (AE), denoted as \textbf{Basel+AE};
\item The above configuration plus dynamic graph learning (DG), denoted as \textbf{Basel+AE+DG};
\item The above configuration plus two-step prediction (TP) introduced in \ref{sec:twostep}, which is the full model denoted as \textbf{Basel+AE+DGL+TP} or \textbf{DST-SCNN}. 
\end{itemize}

\begin{figure}[th]
  \centering
	\includegraphics[width=0.45\textwidth]{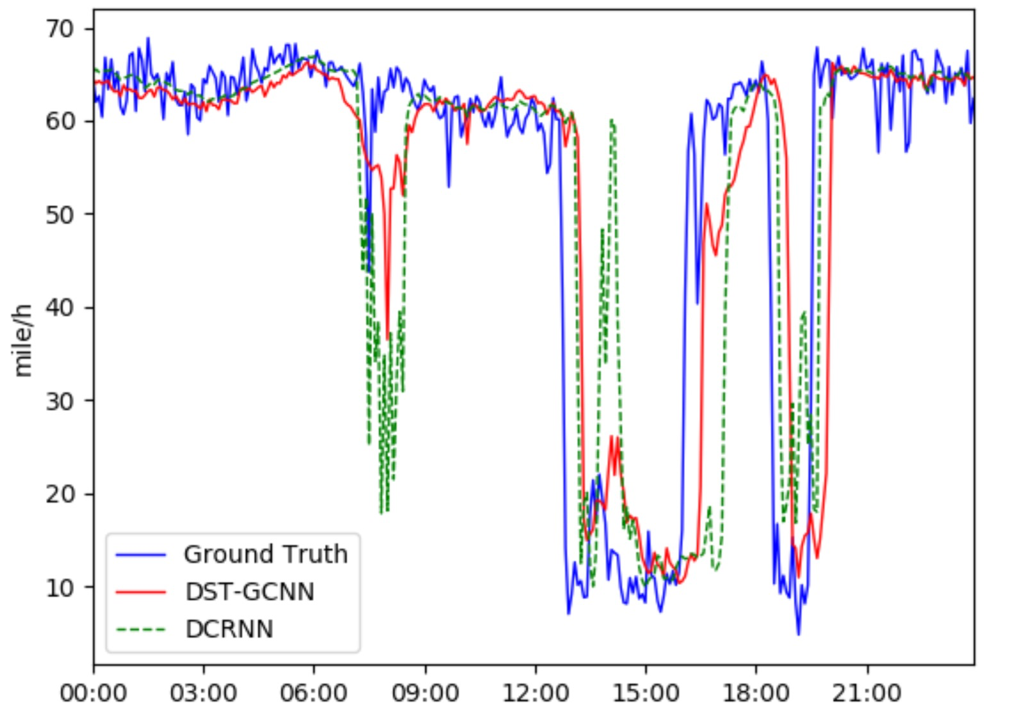}
  \caption{Traffic speed prediction during a day on METR-LA dataset. }
\label{fig:LA_flows}
\end{figure}

The experimental results evaluated on the TaxiBJ test set of all configurations are reported in Table \ref{ablation}. We predict two time steps ahead in all configurations. We can observe that each proposed component consistently reduces the prediction errors and the full model achieves the best performance. The results demonstrate that the auxiliary information embedding, the graph prediction stream and the two-step prediction scheme are all beneficial and complementary to each other. The combination of them accumulates the advantages, therefore achieves the best performance.

%In contrast, LinearSVR gives biased prediction, and FC-LSTM does not catch up with the change of flows.

%The reasons that our method achieves new state-of-the-art are from the following aspects. Compared with traditional methods, our deep model has larger capacity to describe the complex data dependency in traffic network. Second, our method take the dynamic topology of traffic network into consideration while the existing methods don't. As a result, our method can capture the propagation of traffic trends better. Finally, our network is also carefully designed for traffic prediction. The two-step prediction scheme breaks long-term predictions into two short-term predictions and the predictions easier. 
\subsection{Experiments on METR-LA Dataset}

In this subsection, we evaluate the prediction performance of DST-GCNN and the compared methods on METR-LA dataset. We compare DST-GCNN with five different methods, including: 1) Auto-Regressive Integrated Moving Average (ARIMA), which is a well-known method for time-series data forecasting and is widely used in traffic prediction; %2) Linear Support Vector Regression (SVR)(Pedregosa et al. 2011). In order to make use of spatio-temporal data, for each node, we use the historical observations of itself and its neighbors to learn a LinearSVR model;
2) Feed Forward Neural Network (FNN) with two hidden layers and L2 regularization. 
3) Recurrent Neural Network with fully connected LSTM hidden units (FC-LSTM) \cite{sutskever2014sequence}; 
4) DCRNN \cite{li2017diffusion}. This is a recent method which utilizes diffusion convolution and achieves decent results on METR-LA;
5) STGCN \cite{yu2017spatio}. This is a spatio-temporal graph convolutional networks that uses a fixed affinity matrix.

Table \ref{tab:LA_arts} shows the comparison results on METR-LA dataset. For all predicting horizons and all metrics, our method outperforms both traditional statistical approaches and deep learning based approaches. This demonstrates the consistency of our method’s performance for both short-term and long-term prediction.

In Figure \ref{fig:LA_flows}, we also show the qualitative comparison of prediction in a day on the METR-LA dataset. It shows that DST-GCNN can capture the trends of morning peak and evening rush hour better. As can be seen, DST-GCNN predicts the start and end of peak hours which are closer to the ground truth. In contrast, DCRNN does not catch up with the change of traffic data.

\subsection{Experiments on TaxiBJ Dataset}
We also compare the proposed methods with the state-of-the-arts on TaxiBJ dataset. The compared methods include: 1) Seasonal ARIMA (SARIMA); 2) vector auto-regression model (VAR); 3) FCCF \cite{10.1145/2996913.2996934}; 4) FC-LSTM \cite{sutskever2014sequence}; and 5) DCRNN \cite{li2017diffusion}. FCCF utilizes both volume data and auxiliary information including time and weather. Note that we follow the experiments in FCCF that only predict volumes in the next step (30 min later), thus the two-step prediction in our model is not applied. The results by FCCF, SARIMA and VAR were reported in \cite{10.1145/2996913.2996934}. Since only the RMSE results are provided for SARIMA, VAR and FCCF, we compare with these three methods in terms of RMSE metric. For FC-LSTM and DCRNN, we use their default experimental settings in the corresponding papers, and the results are compared in terms of RMSE, MAP and MAPE. We show the results in Table \ref{table:res_bj}.

\begin{table*}[ht]
	\centering
		\renewcommand{\arraystretch}{1.1}
	\begin{tabular} {l|c|c|c|c|c|c}
		\hline
		\multirow{2}{*}{Method} & \multicolumn{3}{|c|}{Out Volumes} & \multicolumn{3}{c}{In Volumes} \\
		\cline{2-7}
		& MAE & RMSE & MAPE & MAE & RMSE & MAPE \\
		\hline
		SARIMA  & - & 21.2 & - & - & 18.9 & -  \\
		VAR & - & 15.8 & - & - & 15.8 & -  \\
		FCCF & - & 14.2 & - & - & 14.1 & -  \\
		FC-LSTM & 11.32 & 14.4 & 13.67\% & 11.92 & 15.3 & 17.30\%  \\
		DCRNN & 10.49 & 13.8 & 13.11\% & 10.71 & 14.5 & 14.46\%  \\
		DST-GCNN & \textbf{9.38} & \textbf{12.0} & \textbf{11.9\%} & \textbf{9.3}& \textbf{12.62} & \textbf{13.27\%} \\
		\hline
	\end{tabular}
	%\vspace{0.2cm}
	\caption{Performance comparison of different methods on the TaxiBJ dataset. MAE, RMSE and MAPE(\%) metrics are compared for different predicting horizons. Bold numbers indicate the best results.}
	\label{table:res_bj}
\end{table*}

From Table \ref{table:res_bj}, we can see that the proposed DST-GCNN achieves the best performance. The comparison results suggests that the proposed STC layer combined with the graph prediction stream is very effective at future traffic prediction. Although the two-step prediction strategy is not utilized in the case of predicting one-step ahead, our method still models the spatio-temporal dependency and the dynamic graph structure robustly.

\subsection{Experiments on CD-HW Dataset}
In this subsection, we evaluate the prediction performance of DST-GCNN and the compared methods on CD-HW dataset.
Because deep learning methods have shown better performance than traditional methods, we only compare our methods with DCRNN and STGCN. Table \ref{tab:cd-hw-arts} shows the comparison results. 
% The compared methods including 1) Auto-Regressive Integrated Moving Average (ARIMA), which is a well-known method for time-series data forecasting and is widely used in traffic flow prediction. 2) Linear Support Vector Regression (LinearSVR) \cite{pedregosa2011scikit}. In order to make use of spatio-temporal data, for each node, we use the historical flows of itself and its neighbors to learn a LinearSVR model. 3) Recurrent Neural Network with fully connected LSTM hidden units (FC-LSTM) \cite{sutskever2014sequence}.

\begin{table}[thp]
	\centering
	\renewcommand{\arraystretch}{1.1}
	\caption{Performance comparison of different methods on the CD-HW dataset. MAE, RMSE and MAPE(\%) metrics are compared for different predicting horizons. Bold numbers indicate the best results.}
	\begin{tabular}{c|c|ccc}  
		\hline
		T & \multicolumn{1}{l|}{Metric} &
		\multicolumn{1}{c}{DCRNN} & \multicolumn{1}{c}{STGCN} &
		\multicolumn{1}{c}{DST-GCNN}  \\
		\hline
		\multirow{3}{*}{30 min} & MAE  & 7.92  & 7.76  &  \textbf{6.33}    \\
		& RMSE    & 12.10 & 11.77  &   \textbf{10.18}  \\
		& MAPE    & 12.4 & 12.2  &  \textbf{10.9}   \\
		\hline
		\multirow{3}{*}{60 min} & MAE  & 9.22 & 8.82 &   \textbf{7.91}   \\
		& RMSE    & 14.32 &  13.61  & \textbf{12.32}    \\
		& MAPE    & 16.3  & 15.9  &  \textbf{14.3}   \\
		\hline
	\end{tabular} 
	\label{tab:cd-hw-arts}
\end{table}

We can observe that our method outperforms the recently deep learning based approaches DCRNN and STGCN. The reason is that: our method takes the dynamic topology of traffic network into consideration while the existing methods don’t. As a result, our method can capture the propagation of traffic trends better. Furthermore, our method could avoid error propagation as in DCRNN.

\subsection{Experimental Results analysis}

The reasons that our method achieves new state-of-the-art are from the following aspects. Compared with traditional methods, our deep model has larger capacity to describe the complex data dependency in traffic network. Second, our method takes the dynamic topology of traffic network into consideration while the existing methods don’t. As a result, our method can capture the propagation of traffic trends better. Finally, our network is also carefully designed for traffic prediction. The two-step prediction scheme breaks long-term predictions into two short-term predictions and makes the predictions easier.

\section{Conclusion and Future Work}

In this paper, we propose an effective and efficient framework DST-GCNN that can predict future traffic flows using surveillance videos. DST-GCNN is able to capture both the dynamics and complexity in traffic. The experiments indicate that our method outperforms other state-of-the-art methods. In the future, we plan to apply the framework to other traffic prediction tasks like pedestrian crowd prediction.

\bibliographystyle{IEEEtran}
\bibliography{DST-GCNN.bib}

\begin{thebibliography}{10}
\providecommand{\url}[1]{#1}
\csname url@samestyle\endcsname
\providecommand{\newblock}{\relax}
\providecommand{\bibinfo}[2]{#2}
\providecommand{\BIBentrySTDinterwordspacing}{\spaceskip=0pt\relax}
\providecommand{\BIBentryALTinterwordstretchfactor}{4}
\providecommand{\BIBentryALTinterwordspacing}{\spaceskip=\fontdimen2\font plus
\BIBentryALTinterwordstretchfactor\fontdimen3\font minus
  \fontdimen4\font\relax}
\providecommand{\BIBforeignlanguage}[2]{{%
\expandafter\ifx\csname l@#1\endcsname\relax
\typeout{** WARNING: IEEEtran.bst: No hyphenation pattern has been}%
\typeout{** loaded for the language `#1'. Using the pattern for}%
\typeout{** the default language instead.}%
\else
\language=\csname l@#1\endcsname
\fi
#2}}
\providecommand{\BIBdecl}{\relax}
\BIBdecl

\bibitem{davis1990adaptive}
G.~A. Davis, N.~L. Nihan, M.~M. Hamed, and L.~N. Jacobson, ``Adaptive
  forecasting of freeway traffic congestion,'' \emph{Transportation Research
  Record}, no. 1287, 1990.

\bibitem{williams2003modeling}
B.~M. Williams and L.~A. Hoel, ``Modeling and forecasting vehicular traffic
  flow as a seasonal arima process: Theoretical basis and empirical results,''
  \emph{Journal of Transportation Engineering-asce}, vol. 129, no.~6, pp.
  664--672, 2003.

\bibitem{lv2015traffic}
Y.~Lv, Y.~Duan, W.~Kang, Z.~Li, and F.~Wang, ``Traffic flow prediction with big
  data: A deep learning approach,'' \emph{IEEE Transactions on Intelligent
  Transportation Systems}, vol.~16, no.~2, pp. 865--873, 2015.

\bibitem{huang2014deep}
W.~Huang, G.~Song, and e.~a. Hong, ``Deep architecture for traffic flow
  prediction: Deep belief networks with multitask learning,'' \emph{IEEE
  Transactions on Intelligent Transportation Systems}, vol.~15, no.~5, pp.
  2191--2201, 2014.

\bibitem{dai2017deeptrend}
X.~Dai, R.~Fu, Y.~Lin, L.~Li, and F.-Y. Wang, ``Deeptrend: A deep hierarchical
  neural network for traffic flow prediction,'' \emph{arXiv preprint
  arXiv:1707.03213}, 2017.

\bibitem{zhang2016deep}
J.~Zhang, Y.~Zheng, and D.~Qi, ``Deep spatio-temporal residual networks for
  citywide crowd flows prediction,'' \emph{national conference on artificial
  intelligence}, pp. 1655--1661, 2016.

\bibitem{li2017graph}
Y.~{Li}, R.~{Yu}, C.~{Shahabi}, and Y.~{Liu}, ``Graph convolutional recurrent
  neural network: Data-driven traffic forecasting,'' \emph{arXiv}, 2017.

\bibitem{yu2017spatio}
B.~Yu, H.~Yin, and Z.~Zhu, ``Spatio-temporal graph convolutional neural
  network: A deep learning framework for traffic forecasting,'' \emph{arXiv
  preprint arXiv:1709.04875}, 2017.

\bibitem{larry1995event}
H.~K. Larry, ``Event—based short—term traffic flow prediction model,''
  \emph{Transportation Research Record}, vol. 1510, pp. 125--143, 1995.

\bibitem{vlahogianni2014short}
E.~I. Vlahogianni, M.~G. Karlaftis, and J.~C. Golias, ``Short-term traffic
  forecasting: Where we are and where we’re going,'' \emph{Transportation
  Research Part C: Emerging Technologies}, vol.~43, pp. 3--19, 2014.

\bibitem{hoang2016fccf}
M.~X. Hoang, Y.~Zheng, and A.~K. Singh, ``Fccf: forecasting citywide crowd
  flows based on big data,'' in \emph{Proceedings of the 24th ACM SIGSPATIAL
  International Conference on Advances in Geographic Information
  Systems}.\hskip 1em plus 0.5em minus 0.4em\relax ACM, 2016, p.~6.

\bibitem{jin2007simultaneously}
X.~Jin, Y.~Zhang, and D.~Yao, ``Simultaneously prediction of network traffic
  flow based on pca-svr,'' \emph{Advances in Neural Networks--ISNN 2007}, pp.
  1022--1031, 2007.

\bibitem{leshem2007traffic}
G.~Leshem and Y.~Ritov, ``Traffic flow prediction using adaboost algorithm with
  random forests as a weak learner,'' in \emph{Proceedings of World Academy of
  Science, Engineering and Technology}, vol.~19, 2007, pp. 193--198.

\bibitem{wu2016short}
Y.~Wu and H.~Tan, ``Short-term traffic flow forecasting with spatial-temporal
  correlation in a hybrid deep learning framework,'' \emph{arXiv preprint
  arXiv:1612.01022}, 2016.

\bibitem{li2017diffusion}
Y.~Li, R.~Yu, C.~Shahabi, and Y.~Liu, ``Diffusion convolutional recurrent
  neural network: Data-driven traffic forecasting,'' \emph{arXiv preprint
  arXiv:1707.01926}, 2017.

\bibitem{jagadish2014big}
H.~V. Jagadish, J.~Gehrke, A.~Labrinidis, Y.~Papakonstantinou, J.~M. Patel,
  R.~Ramakrishnan, and C.~Shahabi, ``Big data and its technical challenges,''
  \emph{Communications of The ACM}, vol.~57, no.~7, pp. 86--94, 2014.

\bibitem{liu2016ssd}
W.~Liu, D.~Anguelov, D.~Erhan, and e.~a. Szegedy, ``Ssd: Single shot multibox
  detector,'' in \emph{ECCV}.\hskip 1em plus 0.5em minus 0.4em\relax Springer,
  2016, pp. 21--37.

\bibitem{henriques2015high}
J.~F. Henriques, R.~Caseiro, P.~Martins, and J.~Batista, ``High-speed tracking
  with kernelized correlation filters,'' \emph{TPAMI}, vol.~37, no.~3, pp.
  583--596, 2015.

\bibitem{howard2017mobilenets}
A.~G. Howard, M.~Zhu, B.~Chen, and e.~a. Kalenichenko, ``Mobilenets: Efficient
  convolutional neural networks for mobile vision applications,'' \emph{arXiv
  preprint arXiv:1704.04861}, 2017.

\bibitem{henaff2015deep}
M.~Henaff, J.~Bruna, and Y.~LeCun, ``Deep convolutional networks on
  graph-structured data,'' \emph{arXiv preprint arXiv:1506.05163}, 2015.

\bibitem{defferrard2016convolutional}
M.~Defferrard, X.~Bresson, and P.~Vandergheynst, ``Convolutional neural
  networks on graphs with fast localized spectral filtering,'' in
  \emph{Advances in Neural Information Processing Systems}, 2016, pp.
  3844--3852.

\bibitem{hammond2011wavelets}
D.~K. Hammond, P.~Vandergheynst, and R.~Gribonval, ``Wavelets on graphs via
  spectral graph theory,'' \emph{Applied and Computational Harmonic Analysis},
  vol.~30, no.~2, pp. 129--150, 2011.

\bibitem{jain2016structural}
A.~Jain, A.~R. Zamir, S.~Savarese, and A.~Saxena, ``Structural-rnn: Deep
  learning on spatio-temporal graphs,'' in \emph{CVPR}, 2016, pp. 5308--5317.

\bibitem{sutskever2014sequence}
I.~Sutskever, O.~Vinyals, and Q.~V. Le, ``Sequence to sequence learning with
  neural networks,'' in \emph{Advances in neural information processing
  systems}, 2014, pp. 3104--3112.

\bibitem{10.1145/2996913.2996934}
\BIBentryALTinterwordspacing
M.~X. Hoang, Y.~Zheng, and A.~K. Singh, ``Fccf: Forecasting citywide crowd
  flows based on big data,'' in \emph{Proceedings of the 24th ACM SIGSPATIAL
  International Conference on Advances in Geographic Information Systems}, ser.
  SIGSPACIAL ’16.\hskip 1em plus 0.5em minus 0.4em\relax New York, NY, USA:
  Association for Computing Machinery, 2016. [Online]. Available:
  \url{https://doi.org/10.1145/2996913.2996934}
\BIBentrySTDinterwordspacing

\end{thebibliography}
%\begin{thebibliography}%{00}
%\bibliography{reference.bib}
%\end{thebibliography}

\EOD

\end{document}